\documentclass[10pt,journal,compsoc]{IEEEtran}

\ifCLASSOPTIONcompsoc
  \usepackage[nocompress]{cite}
\else
  \usepackage{cite}
\fi

\usepackage{url}
\usepackage[utf8]{inputenc} 
\usepackage[T1]{fontenc}    
\usepackage{booktabs}       
\usepackage{amsfonts}       
\usepackage{amsthm}
\usepackage{nicefrac}       
\usepackage{microtype}      
\usepackage{xcolor}         
\usepackage{amsmath}
\usepackage{subfigure}
\usepackage{graphicx}
\usepackage{enumitem}
\usepackage{mathtools}
\usepackage{wrapfig}
\usepackage{algorithm}
\usepackage{algorithmic}
\usepackage{multirow}

\newtheorem{definition}{Definition}

\newtheorem{proposition}{Proposition}
\newtheorem{theorem}{Theorem}

\usepackage{comment}
\usepackage{bm}
\usepackage{makecell}
\usepackage{bbm}
\usepackage{hyperref}

\usepackage{tikz}

\usepackage[utf8]{inputenc}

\renewcommand{\algorithmicrequire}{\textbf{Input:}}
\renewcommand{\algorithmicensure}{\textbf{Output:}}
\def\cpar{\hss\egroup\line\bgroup\hss}

\begin{document}
%
\title{Alpha and Prejudice: Improving $\alpha$-sized Worst-case Fairness via Intrinsic Reweighting}

\author{Jing Li, 
        Yinghua Yao,
        Yuangang Pan,
        Xuanqian Wang,
        Ivor W. Tsang~\IEEEmembership{Fellow,~IEEE},
        and~Xiuju Fu
\IEEEcompsocitemizethanks{
\IEEEcompsocthanksitem Jing~Li, Yinghua~Yao, Yuangang~Pan, and Ivor~W.~Tsang are with the Institute of High Performance Computing, Agency for Science, Technology and Research, Singapore, and also with the
Centre for Frontier AI Research, Agency for Science, Technology and Research, Singapore. \protect\\
E-mail: \{kyle.jingli, eva.yh.yao, yuangang.pan, ivor.tsang\}@gmail.com
\IEEEcompsocthanksitem Xuanqian~Wang is with School of Computer Science, Beihang University, Beijing, China.\protect\\
Email: wwxxqq@buaa.edu.cn
\IEEEcompsocthanksitem Xiuju~Fu is with the Institute of High Performance Computing, Agency for Science, Technology and Research, Singapore.\protect\\
Email: fuxj@ihpc.a-star.edu.sg.
}}


\markboth{Journal of \LaTeX\ Class Files,~Vol.~xx, No.~x, August~2022}%
{Shell \MakeLowercase{\textit{et al.}}: Bare Demo of IEEEtran.cls for Computer Society Journals}

\IEEEtitleabstractindextext{%
\begin{abstract}
Worst-case fairness with off-the-shelf demographics achieves group parity by maximizing the model utility of the worst-off group. Nevertheless, demographic information is often unavailable in practical scenarios, which impedes the use of such a direct max-min formulation. Recent advances have reframed this learning problem by introducing the lower bound of minimal partition ratio, denoted as $\alpha$, as side information, referred to as ``$\alpha$-sized worst-case fairness'' in this paper. We first justify the practical significance of this setting by presenting noteworthy evidence from the data privacy perspective, which has been overlooked by existing research. Without imposing specific requirements on loss functions, we propose reweighting the training samples based on their intrinsic importance to fairness. Given the global nature of the worst-case formulation, we further develop a stochastic learning scheme to simplify the training process without compromising model performance. Additionally, we address the issue of outliers and provide a robust variant to handle potential outliers during model training. Our theoretical analysis and experimental observations reveal the connections between the proposed approaches and existing ``fairness-through-reweighting'' studies, with extensive experimental results on fairness benchmarks demonstrating the superiority of our methods.  

\end{abstract}


\begin{IEEEkeywords}
Fairness learning, worst-case formulation, reweighted samples, stochastic optimization, outliers.
\end{IEEEkeywords}}

\maketitle

\IEEEdisplaynontitleabstractindextext

%
\IEEEpeerreviewmaketitle

\IEEEraisesectionheading{\section{Introduction}\label{sec:introduction}}
\IEEEPARstart{N}{umerous} sectors within society have been identified as revealing instances of prejudice across demographic groups, such as credit offerings~\cite{steel2010web}, recidivism prediction~\cite{dressel2018accuracy}, allocation of health-care resources~\cite{dieleman2021us}, and more. As AI systems become extensively integrated into contemporary decision-making processes, concerns regarding discrimination hold significance in the realm of machine learning model development~\cite{ntoutsi2020bias}. With prepared demographic information, group fairness~\cite{zafar2017fairness,zafar2017fairnessWWW,DBLP:conf/innovations/KleinbergMR17,agarwal2018reductions}, a.k.a. statistical fairness~\cite{carey2023statistical}, has been established, which typically restricts the trained model to behave consistently across different groups. Although stated in an in-processing manner, group fairness achieved through pre-processing~\cite{kamiran2012data,zemel2013learning,feldman2015certifying,xie2017controllable,zhang2022review} and post-processing~\cite{hardt2016equality,DBLP:conf/aaai/PutzelL22,jang2022group} can be also interpreted in a similar fashion; the former requires the learned representation of each group to be statistically aligned with each other and the latter treats the model after standard training as a black box and focuses on how to barely adjust model outputs to remove group variations. 

In the case of a multi-valued sensitive attribute, such as race or religion, a naive approach to achieving fairness can be characterized by minimizing the summed discrimination across all pairs of groups. Given a predefined model utility, John Rawls' theory~\cite{rawls2001justice} promotes maximizing the utility (e.g., classification accuracy) of the worst-off group, thereby ensuring the consent of minorities. Because if a model's capacity is limited, encouraging the utility of worst-off group at each iteration will eventually reduce the group disparity. Following the convention of~\cite{martinez2021blind}, we use \emph{worst-case fairness} to refer to this type of fairness formulations. Note that worst-case fairness is distinct from notions like disparate treatment or disparate impact~\cite{barocas2016big}, which focuses on decisions. Instead, it aligns more closely with utility-based fairness research, such as fair PCA~\cite{samadi2018price, DBLP:conf/iclr/VuTYN22} or fair regression~\cite{chi2021understanding}, where the group disparity on reconstruction errors or regression errors is considered.

Recent developments in worst-case fairness focus on the scenarios where demographic information is not available during the training process due to laws or privacy in practice~\cite{veale2017fairer,holstein2019improving}. This restriction introduces a new challenge for training fair models, as we are no longer have access to group information, which was previously accessible through data annotations. Some works propose using proxy attributes, either by selecting observed ones~\cite{yan2020fair,zhu2023weak} or predicting from data~\cite{DBLP:conf/ijcai/GrariLD22,belli2022county}. We follow the assertion of~\cite{martinez2021blind} that an interested attribute could arbitrarily partition the training data, implying that proxy attributes may not be sufficient in this case. To still enable a worst-case formulation without demographics, they propose that group size should be constrained by a parameter $\alpha$, as has been argued in distributional robust optimization~\cite{duchi2021learning}. Although starting from the necessity of modeling, they pave the way for learning worst-case fairness with group prior used during model training only, followed by numerous subsequent works~\cite{zhai2021doro,zeng2022outlier,chai2022self}.  

We believe there is a need to explain why it is possible to obtain prior information about group size even when demographics are too sensitive to acquire beforehand. In this paper, we take the view of data privacy~\cite{dwork2014algorithmic}, offering appropriate evidence for this research topic. On the other hand, by reviewing existing methods we realize that the solutions of achieving worst-case fairness can be further improved, because the state-of-the-art methods transform the population-based formulation into greedily upweighting the higher-loss samples instance-wise\footnote{This is also understood as an over-fitting issue~\cite{martinez2021blind}, which they mitigate by adopting cross-validation.}, which however overlooks the true contribution of each example and thus fails to obtain satisfactory model performance. Our assertion stems from influence function~\cite{koh2017understanding}, which reveals that sample importance can be characterized from individual gradients rather than their loss values only. Interestingly, we realized that the individual gradients used to compute sample importance provide a valuable representation that could be applied to address the outliers problem, which has been identified as one of the key risks in worst-case formulations.


%
The main contributions of this work can be summarized as follows.
\begin{itemize}
    \item We present supporting evidence grounded in local differential privacy~\cite{dwork2014algorithmic} that justifies the practice of \emph{$\alpha$-sized worst-case fairness} where demographic information in the training data is unavailable, but side information regarding the minimal partition size is accessible. 
    \item We propose the Intrinsic ReWeighting (IRW) method to address $\alpha$-sized worst-case fairness, reweighting each training sample by computing its individual contribution to improving the utility of the worst-off group. IRW is a gradient-based approach that provides a more reliable reweighting compared to existing methods, which heuristically assign larger weights to samples with higher losses.
    \item We enhance the efficiency of IRW by incorporating stochastic updates which overcome the limitations imposed by the global nature of worst-case formulations. Additionally, we extend IRW to handle scenarios with outliers by leveraging individual gradients as representations, allowing the method to effectively identify and remove potential outliers.
    \item We demonstrate the superiority of IRW through experimental comparisons with existing methods across various datasets. Our experiments also reveal the connections of all methods from the per-sample reweight strategy and exploit the potential impact of different choices of evaluation metrics. 
\end{itemize}

\section{Related Work}\label{sec:related_work}
Despite the vast body of literature on fairness learning, we revisit three branches of research that are closely related to our work.


\textbf{Utility-centric Fairness.} 
Following Rawlsian fairness~\cite{rawls2001justice}, many fairness learning works~\cite{hashimoto2018fairness,heidari2019moral,lahoti2020fairness,mitchell2021algorithmic,gupta2024fairly} have concentrated on model's utility across different groups. In classification tasks, the accuracy of each group data is expected to be same with each other, also termed as subgroup robustness~\cite{martinez2021blind,liu2021just}. Regarding regression task, the utility is typically characterized by regression error~\cite{DBLP:conf/iclr/0002CAG20}, while in PCA the utility can be described into reconstruction error~\cite{samadi2018price,DBLP:conf/iclr/VuTYN22}. With utility in consideration, fairness can be achieved by minimizing utility discrepancies across groups. Recent work~\cite{martinez2020minimax} treats learning on every group as a multiple coupled objective, and then seeks a set of optimal classifier parameters along the Pareto front that best balances the utility among all groups. Our work falls in the accuracy-based fair classification tasks but also consider alternative evaluations metrics, such as F1 score for imbalanced datasets.

\textbf{Fairness without Demographics.} 
A direct workaround for this problem is using proxy attributes. One approach is to perform clustering on observed features and treat cluster indicators as groups~\cite{chaudhary2023practical}. Another is to select observed attributes under the assumption that these attributes are usually correlated~\cite{yan2020fair}. Alternatively, one might approximate the sensitive attributes using a defined causal graph~\cite{DBLP:conf/ijcai/GrariLD22} or employ proxy models for prediction~\cite{belli2022county}. Given a group size, the Rawlsian fairness can be approximated without need of demographics, such as in~\cite{hashimoto2018fairness,duchi2021learning}. The key insight behind is that true worst-off group utility can be lower bounded when group size prior is known. We emphasize that such an approximation is dependent on the utility-centric fairness.

\textbf{Sample Reweighting.}
Reweighting~\cite{calders2009building} is one of the earliest strategies to mitigate bias. However, standard reweighting techniques are designed for cohort-based approaches~\cite{hajian2012methodology,zemel2013learning}, making them inapplicable to demographics-unaware scenarios. A pioneering work that connects worst-case fairness and sample reweighting is~\cite{hashimoto2018fairness}, which observes that the samples with higher loss tend to receive more focus during training. This finding inspires the subsequent research~\cite{lahoti2020fairness,martinez2021blind} to optimize learnable sample weights. Our work builds on this insight by evaluating each sample's importance through its influence on the fairness objective. However, unlike~\cite{chai2022self}, our method is not dependent on any additional validation set with demographic annotation, thereby being free of additional requirement on training dataset. 

\section{Problem Formulation}
\subsection{Worst-case Fairness}
Let predictive accuracy serve as the utility of a classification model $f_{\theta}$ parameterized by $\theta$. If a sensitive attribute splits training data into $K$ groups, i.e., $G = \{G_1, G_2,...,G_K\}$, Rawlsian max-min fairness, a.k.a. the worst-case fairness~\cite{martinez2021blind}, optimizes the following problem,
\begin{equation}\label{eq:utility_as_acc}
    \theta^* = \arg\max_{\theta}\min_{k\in [K]} \text{ACC}_k(\theta),
\end{equation}
where $\text{ACC}_k(\theta)$ represents the accuracy of classifier $f_{\theta}$ on group $G_k$, i.e., $\text{ACC}_k(\theta) =\frac{1}{|G_k|}\sum_{i=1}^{|G_k|}\mathbbm{1}({f_{\theta}(x)= y })$. Let worst-off group accuracy $\text{ACC}_{wg}(\theta) = \min_{k \in [K]} \text{ACC}_k(\theta)$. As $\text{ACC}_k(\theta) \le 1 (\forall k \in [K])$, we have
\begin{equation}\label{eq:fairness_worst_group_acc}
|\text{ACC}_i(\theta) - \text{ACC}_j(\theta)| \le 1-\text{ACC}_{wg}(\theta), \;\; \forall i,j \in [K]
\end{equation}
which indicates the utility difference between any two groups are upper bounded by $1-\text{ACC}_{wg}(\theta)$. With this insight, Eq.~\eqref{eq:utility_as_acc} optimizes $\theta$ to maximize $\text{ACC}_{wg}(\theta)$ iteratively, thereby reducing the utility gap among groups.

To avert the non-differentiable indicator function $\mathbbm{1}(\cdot)$ during optimization, Eq.~\eqref{eq:utility_as_acc} is practically converted to minimizing the expected risk over the worst-off group,
\begin{equation}\label{eq:minmax_group_loss}
    \theta^* =\arg\min_{\theta}\max_{k\in [K]} \mathcal{J}_k(\theta), \;\; \mathcal{J}_k(\theta)=\mathbb{E}_{z\sim P_k}[\ell(\theta;z)]
\end{equation}
where the variable $z =(x,y)$, $P_k$ denotes the distribution from which the members of $G_k$ are drawn, and $\ell(;)$ is any feasible classification loss function, e.g., cross entropy. Similar to $\text{ACC}_{wg}(\theta)$, we denote $\mathcal{J}_{wg}(\theta) = \max_{k\in[K]} \mathcal{J}_k(\theta)$ for simple statement later. In particular, if $f_{\theta}$ belongs to the hypothesis class which achieves Pareto optimal classifiers, Eq.~\eqref{eq:minmax_group_loss} is able to achieve the well-known harmless fairness if a binary sensitive attribute is applied. While for multi-value sensitive attributes, it has been found difficult to find an equal-group-loss plane~\cite{martinez2020minimax}. 

When demographic information is unknown, group partition $G$ then cannot not be accessible, making the inner optimization of Eq.~\eqref{eq:minmax_group_loss} intractable. Section~\ref{sec:related_work} has revisited how existing studies remedy this issue. Although not all particularly devised for the worst-case fairness, they are potentially used as workarounds if corresponding conditions are provided.

\subsection{Setting: $\alpha$-sized Worst-case Fairness}
\subsubsection{Minimal Group Proportion for Bounding}\label{sec:group_proportion_bounding}
Distributional robustness around the empirical distribution with radius shrinking~\cite{namkoong2017variance} offers a bridge between group and global empirical risks. Inspired by this finding, the pioneer work~\cite{hashimoto2018fairness} addresses worst-case fairness without demographics by introducing the side information about group partitions. Specifically, they presume a lower bound on the group proportions $\alpha \le \min_{k\in[K]} \frac{|G_k|}{|G|}$ by which a surrogate objective can be formulated. Formally, it minimizes the expected risk over a perturbed distribution, i.e., $Q$
\begin{equation}\label{eq:distributionl_robust_chi_square_ball}
\mathcal{J}_{dr}(\theta;\alpha):= \sup_{Q\in \mathcal{B}(P,r)}\mathbb{E}_{z\sim Q} [\ell(\theta;z)],
\end{equation}
where $\mathcal{B}(P,r)$ is chi-squared ball around the original data distribution $P$ of radius $r$ being $(1/\alpha-1)^2$. $\mathcal{J}_{dr}(\theta;\alpha)$ is provably to bound the risk of each group $\mathcal{J}_k(\theta)$, and hence the worst-off group risk $\mathcal{J}_{wg}(\theta)$. In fact, given the minimal proportion size $\alpha$ and the total number of training examples $N$, it is verified that the empirical risk (denoted by adding a $\hat{}$ for differentiation, and same to others) of the worst-off group $\hat{\mathcal{J}}_{wg}(\theta)$ is upper bounded by the average risk of examples which achieve top $N\alpha$ largest losses
\begin{equation}\label{eq:bounded_by_top_N_alpha}
    \begin{aligned}
    \hat{\mathcal{J}}_{wg}(\theta)
    &= \max_{k\in[K]} \frac{1}{|G_k|}\sum_{i=1}^{|G_k|}\ell(\theta;z_i) \\
    &\le  \frac{1}{N\alpha} \sum_{i=1}^{N\alpha} \ell(\theta;z^{(i)}):=\hat{\mathcal{J}}_{N\alpha}(\theta;\alpha)
    \end{aligned}
\end{equation}
where $\ell(\theta;z^{(i)})$ denotes $i$-th largest loss yielded by $z^{(i)}$, and $\hat{\mathcal{J}}_{N\alpha}(\theta;\alpha)$ is defined as the empirical risk on the group whose members have largest $N\alpha$ losses. Although the inequality of Eq.~\eqref{eq:bounded_by_top_N_alpha} is easy to reach, due to the nature of $\mathcal{J}_{dr}(\theta;\alpha)$, we instantly have the following proposition.
\begin{proposition}\label{prop1}
Given a bound $\alpha$ for the minimal group proportion, we have $\mathcal{J}_{wg}(\theta) \le \mathcal{J}_{N\alpha}(\theta;\alpha) \le \mathcal{J}_{dr}(\theta;\alpha)$ for all $\theta \in \Theta$. 
\end{proposition}
\noindent The proof of Proposition~\ref{prop1} is left to Appendix. Although optimizing $\mathcal{J}_{dr}(\theta;\alpha)$ leads to a convex formulation~\cite{namkoong2017variance}, we focus on minimizing $\hat{\mathcal{J}}_{N\alpha}(\theta;\alpha)$ of Eq.~\eqref{eq:bounded_by_top_N_alpha} that is more efficient to improve the worst-off group performance. Now, we formally frame the definition of the problem setting.
\begin{definition}
\textbf{($\alpha$-sized Worst-case Fairness)} A classifier $f_{\theta}$ is said to satisfy $\alpha$-sized worst-case fairness principle on a training set of $N$ examples if it maximizes the utility of the subgroup $S$, denoted by $U_S$, which is composed of $N\alpha$ examples and is with the lowest utility. 
\begin{equation}\label{eq:definition_alpha_worst_case_fairness}
\theta^* = \arg\max_{\theta \in \Theta} \min_{|S|=N\alpha} U_{S}(\theta).
\end{equation}
\end{definition}
\noindent Similar to Eq.~\eqref{eq:minmax_group_loss}, by taking predictive accuracy as utility and selecting $N\alpha$ examples with largest losses, the inner objective of Eq.~\eqref{eq:definition_alpha_worst_case_fairness} can be implemented by the negative $\hat{\mathcal{J}}_{N\alpha}(\theta;\alpha)$. Hence, the full objective is written as
\begin{equation}\label{eq:minimize_top_N_alpha_loss}
\min_\theta \frac{1}{N\alpha} \sum_{i=1}^{N\alpha} \ell(\theta;z^{(i)}).
\end{equation}
Note that $z^{(i)} (i=1,2,...,N\alpha)$ is dependent on $\theta$.

\subsubsection{Deriving $\alpha$ from Privacy Perspective}
Note that both $\mathcal{J}_{dr}(\theta;\alpha)$ and $\mathcal{J}_{N\alpha}(\theta;\alpha)$ introduce group proportion bound $\alpha$ in a mathematical manner. Recall that demographics are not available often because of privacy concerns. To enhance data privacy, we show in the following example that $\alpha$ could be properly estimated by randomized response~\cite{kasiviswanathan2011can} even without true attribute labeling.

\noindent \textbf{Example.} Suppose gender is considered a sensitive attribute, with ``female'' being the protected group. When collecting data, we instruct each participant to perform a coin flip. If tails, they should respond truthfully. If heads, they should flip a second coin and respond ``female'' if heads and ``male'' if tails. We denote the true female proportion and the frequency of ``female'' answers by $\alpha$ and $\beta$, respectively. We know that the expected number of ``female'' answers is $1/4$ times the number of participants who are male plus $3/4$ times the actual number of female participants, that is, $\beta=(1/4)(1-\alpha)+(3/4)\alpha$. Therefore, we can estimate the female proportion by $\alpha = 2\beta-1/2$. 

This example shows that the collected answers about gender can be very noisy, as each individual has a plausible deniability to their outcomes, from which privacy comes. From the theory of differential privacy, we have the following theorem.
\begin{theorem}
The randomized response described in the above example is $(\ln3,0)$-differentially private.
\end{theorem}
\noindent The proof can be referred to Chapter 3 of~\cite{dwork2014algorithmic}. Note that the probability of deniability can be changeable if one is apt to adopt some other strategies. Generally, the estimated $\alpha$ will become more accurate if the number of participants is larger. And we use this $\alpha$ as an approximation of the minimal group proportion. Regarding a multi-value attribute, e.g., race, if we are aware which group is to be protected, we can follow this example by simply treating them as a binary case, e.g., ``black'' and ``non-black''.

\section{Reweighting Solution}
\begin{figure}
\centering
\includegraphics[width=0.5\textwidth]{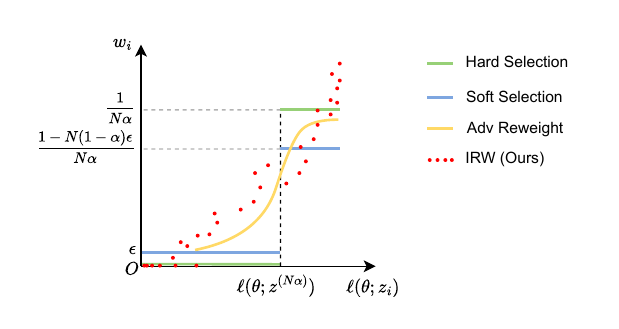}
\caption{Comparison of four weights learning strategies. The weight of our IRW method is not solely determined by loss; therefore we use discrete scatter plots to represent the variance introduced by other factors, i.e., gradients. \label{fig:motivation}}
\end{figure}
\subsection{Per-sample Weight Learning}\label{sec:IRW}
Optimizing Eq.~\eqref{eq:minimize_top_N_alpha_loss} is also referred as minimizing top-${N\alpha}$ loss in the recent study~\cite{chai2022self}, which essentially encourages the model to fit more difficult examples only, that is,
\begin{equation}\label{eq:top_N_alpha_form_2}
\min_{\theta} \frac{1}{N\alpha}\sum_{i=1}^{N}[\ell(\theta;z_i)-\ell(\theta;z^{(N\alpha+1)})]_+ + \ell(\theta;z^{(N\alpha+1)}),
\end{equation}
where $[a]_+ = \max(a,0)$. Note that this objective minimizes the $\alpha$-fraction of training points that have the highest loss~\cite{liu2021just}. If $\alpha$ is small, there will be very few examples that participate the training per iteration, incurring an inferior model utility (See the results of CVaR~\cite{duchi2021learning} in Table~\ref{tbl:specific_attributes}). 
Essentially, this sample selection process can be understood as assigning a set of weights to the training examples, where Eq.~\eqref{eq:top_N_alpha_form_2} effectively performs a hard selection. To better utilize the training samples, a soft selection approach can be adopted by maintaining a minimum weight for samples whose losses are below the threshold. Since the weights lie on a $(N-1)$-dimensional simplex, denoted as $\Delta^{N-1}$, we can write the per-sample weight as
\begin{eqnarray}\label{eq:top_N_alpha_distribution}
w_i=\begin{cases} \epsilon, & \text{if} \; \ell(\theta;z_i) < \ell(\theta;z^{(N\alpha)}) \\
\frac{1-N(1-\alpha)\epsilon}{N\alpha}, &   \text{otherwise.} \end{cases}
\end{eqnarray}
Obviously, Eq.~\eqref{eq:top_N_alpha_distribution} reduces to the hard selection if $\epsilon=0$. Instead of manually assigning binary weights, adversarial reweighting fairness works~\cite{lahoti2020fairness,martinez2021blind} propose that these weights are learnable through the following form:
\begin{equation}\label{eq:adv_weight_obj}
\max_{w \in \Delta^{N-1}} \sum_{i=1}^{N} w_i\ell(\theta;z_i) \;\; w \in C,
\end{equation}
where $C$ denotes any predefined constraint which could be a specific bound for weights~\cite{martinez2021blind} or a complexity control applied to the function that generates weights~\cite{lahoti2020fairness}, preventing from trivial solutions to Eq.~\eqref{eq:adv_weight_obj}.
\begin{theorem}\label{theorem_2}
If the feasible set defined by $C$ is convex, the optimal solution of Eq.~\eqref{eq:adv_weight_obj} implies that a higher loss corresponds strictly to a higher weight.
\end{theorem}
\noindent The proof of Theorem~\ref{theorem_2} is left to Appendix. Theorem~\ref{theorem_2} suggests that the fairness methods which optimize Eq.~\eqref{eq:adv_weight_obj} are essentially searching for a set of weights that are non-monotonic with respect to the loss, guided by a prior belief encoded in $C$. We summarize these weight distributions into Fig.~\ref{fig:motivation}, where adversarial reweighting can be depicted as a smooth function of loss. Following the formulation of~\cite{martinez2021blind}, the minimum weight $\epsilon$ is maintained for the samples whose losses are relative smaller.

\subsection{Intrinsic Reweighting (IRW)}
The weight of each training sample should reflect its actual contribution to the fairness objective. We challenge the strategy that samples with larger losses should always be assigned larger weights. Because a sample with a large loss but a zero gradient cannot be considered contributive. For example, this can occur when inputs are negative in ReLU activation. Please also skip to Fig.~\ref{fig:weight_distribution} for experimental justifications. 

With this understanding in mind, we look back to Eq.~\eqref{eq:minimize_top_N_alpha_loss} and consider reassigning the weights to the training examples, which can directly reduce the largest $N\alpha$ losses. That means, we intend to learn a weight assignment such that by minimizing reweighted losses we can also achieve the $\alpha$-sized worst-case fairness: 
\begin{subequations}
\begin{align}
    &\theta^*(w) = \arg\min_{\theta} \sum_{i=1}^N w_i\ell(\theta;z_i), \label{eq:reweighted_loss}\\
    &w^* = \arg\min_w \underbrace{\frac{1}{N\alpha}\sum_{i=1}^N [\ell(\theta^*(w);z_i)-\ell(\theta^*(w);z^{(N\alpha+1)})]_+}_{\mathcal{F}(\theta^*(w))}. \label{eq:weights_for_fairness}
\end{align}
\end{subequations}
Finding the optimal $w^*$ in Eq.~\eqref{eq:weights_for_fairness} is difficult because we need to first optimize Eq.~\eqref{eq:reweighted_loss}. Inspired by the idea of examples reweighing via influence function~\cite{ren2018learning}, for each example $z_i$ we approximate its weight at $t$-th iteration by first perturbing its weight by $\epsilon_{t,i}$ and taking the single step gradient evaluated at zero
\begin{equation}\label{eq:gradient_similarity}
v_{t,i} = \frac{\partial \mathcal{F}(\theta_t)}{\partial \epsilon_{t,i}} \Bigg |_{\epsilon_{t,i}=0}
\approx (\nabla_\theta \mathcal{F}(\theta_t))^\top \nabla_{\theta} \ell(\theta_t;z_i).
\end{equation}
Without reiterating the details of how the approximation is derived, we can observe that the contribution of each sample depends on its gradient similarity with $\nabla_\theta \mathcal{F}(\theta_t)$, the gradient of worst-off group. To avoid the negative weights, we further rectify each $v_{t,i}$ and normalize the weights,
\begin{equation}
\label{eq:per_sample_weight}
w_{t,i} = \frac{[v_{t,i}]_+} {\sum_j [v_{t,j}]_+ + \delta\left(\sum_j [v_{t,j}]_+\right)},
\end{equation}
where $\delta(a)=1$ if $a=0$ and otherwise $\delta(a)=0$, which rules out a zero denominator. 

We can see from Eq.~\eqref{eq:gradient_similarity}, the value of $v_{t,i}$ is determined by the gradient similarity between $i$-th training example and the top $N\alpha$ largest training samples. Intuitively,
an example whose loss is above $\ell(\theta_t;z^{(N\alpha+1)})$ will be more likely selected as a component of $\mathcal{F}(\theta_t)$ and thus the inner product tends to be positive. This agrees with the existing loss-based reweighting principle. In particular, we highlight that the samples whose loss are below $\ell(\theta_t;z^{(N\alpha+1)})$ also have chance to produce a positive $v_{t,i}$ if its direction is close to the combined gradient $\nabla_\theta \mathcal{F}(\theta_t)$. In Section~\ref{sec:per-sample_reweight_mechanism}, we will examine how the per-sample weights obtained by IRW disrupt the loss-based ordering. 

\subsection{Stochastic Update}
\label{sec:stochastic_update}
Please note that Eqs.~\eqref{eq:reweighted_loss} and~\eqref{eq:weights_for_fairness} are both defined globally, meaning that we need to optimize model parameters and sample weight with full-batch data update. To scale up to large datasets, we adopt stochastic update by proposing two schemes. (i)~\emph{Global.} At the beginning of each epoch, we identify the top-$N\alpha$ samples by Eq.~\eqref{eq:weights_for_fairness} and compute sample weights $w^*$ across full training data. By assuming that the weights change within each epoch can be negligible, as the same style as~\cite{martinez2021blind}, we do not re-compute $w^*$ for them to save computation costs. (ii)~\emph{Local.} We use local top-$B\alpha$ samples on each mini-batch for Eq.~\eqref{eq:weights_for_fairness}, and compute batch-wise sample weights on a simplex of $\Delta^{B-1}$. The difference of weight scale between two schemes will be absorbed into learning rates.

\textbf{Complexity.} The additional cost of IRW with respect to a standard training is the instance-wise backward propagations, as shown by step 5 in Algorithm~\ref{alg:1}. As is known to us, modern neural networks stores cumulative statistics in a batch rather than individual sample's gradients. A naive solution is to perform single-sample mini-batches, ideally in parallel, which remains too slow and incompatible with batch-based operators like batch normalization. We remedy this problem by using function transforms~\footnote{\url{https://pytorch.org/tutorials/intermediate/per_sample_grads.html}} which compute per-sample-gradients in an efficient way. Given this implementation, we realize that the time complexity difference between above two training schemes primarily arises from the sorting process. For the global weights in scheme~(i), finding the top $N\alpha$ from $N$ training examples based on their losses has a time complexity of $O(N\log (N{\alpha}))$. In contrast, for the local weights of scheme~(ii), the time-complexity is $O(N\log(B\alpha))$. In practical implementation, we experimentally observe that scheme~(i) is slightly faster due to the multiple graph retention steps in scheme~(ii) (See Section~\ref{sec:globla_vs_local}).

\section{Outliers Removal from Gradient View}\label{sec:outliers_removal}
The worst-case formulation, i.e., Eq~\eqref{eq:minimize_top_N_alpha_loss}, is vulnerable to outliers. Since we have the gradient for each training sample, we can filter out outliers by naturally utilizing such gradient information, which is more reliable than loss values, as we have claimed Section~\ref{sec:IRW}. 

Inspired by ~\cite{zeng2022outlier}, we develop a variant of IRW, named by IRWO, to handle potential outliers during training. Specifically, we first compute pairwise distance between every two samples' gradient within a batch. Let us define
$g_i := \nabla_{\theta} \ell(\theta;z_i)$ where we omit the iteration index for $\theta$ for convenience. Then each entry of the distance matrix $M$ is derived by 
\begin{equation}\label{eq:distance_matrix}
M_{ij} = 1- \frac{\tilde{g_i}^{\top} \tilde{g_j}}{||\tilde{g_i}|| \cdot||\tilde{g_j}||},
\end{equation}
where the decentralized gradient $\tilde{g_i} = g_i-\bar{g}$ with $\bar{g}$ as the mean of per-sample gradient. We treat each row of the distance matrix $M$ as the new representation of each sample, which better characterizes the local density and also improves the clustering efficiency when parameter space is larger than data size $|\theta|>|\mathcal{D}|$. We then identify outliers by searching for low-density regions. In our method, we adopt DBSCAN clustering\footnote{\url{https://scikit-learn.org/stable/modules/generated/sklearn.cluster.DBSCAN.html}} before selecting the worst-off group. Typically, training samples labeled as $-1$ after clustering are considered outliers. Since clustering are conducted on entire training set\footnote{Performing clustering on a mini-batch may lead to misidentifying minority samples as outliers.}, which remains expensive to acquire outliers, we choose to periodically perform this process. For instance, we can do every $\tau$ steps, as presented in summarized Algorithm~\ref{alg:1}, which describes the detailed process of local update scheme. In our experiments, $\tau$ is set as the number of batches, and thus we do outliers removal once per epoch. For the global update, i.e., scheme~(i) of Section~\ref{sec:stochastic_update} which requires to prepare all the sample weights before stochastic optimization, we can simply merge steps 14-17 into outliers removal procedure.

\textbf{Related strategies.} We present different methods for handling outliers in worst-case fairness studies. ARL~\cite{lahoti2020fairness} addresses this problem by employing a liner adversary which helps focus on computationally-identifiable errors. DORO~\cite{zhai2021doro} views a certain proportion (hyper-parameter) of samples which achieve largest losses as outliers, purifying the worst-off group. GraSP~\cite{zeng2022outlier} conducts density-based clustering in gradient space where both outliers and groups are indicated by clustering labels. We follow the idea of GraSP to remove potential outliers before we compute per-sample weight, i.e., Eq.~\eqref{eq:gradient_similarity}, since the well-prepared per-sample gradients are ready to use. The modified method is named by IRWO for simplicity.  Their experimental comparison can be referred to Section~\ref{sec:robustness_to_outliers}.

\begin{algorithm}[t]
\caption{IRW algorithm for $\alpha$-sized worst-case fairness \label{alg:1}}
\begin{algorithmic}[1] 
\renewcommand{\algorithmicrequire}{\textbf{Input:}}
\renewcommand{\algorithmicensure}{\textbf{Output:}}
\REQUIRE Training set $\mathcal{D}=\{z_i\}_{i=1}^N$, where $z_i =(x_i,y_i)$, period $\tau$, and group ratio $\alpha$
\ENSURE Learned model parameterized by $\theta$
\STATE Initialize parameters $\theta$
\FOR{$t=0,1,...,T-1$}
\IF{Outliers exist and $T \bmod\tau=0$}
    \STATE Perform forward passes and get $\{\ell(\theta_t;z_i)\}_{i=1}^{|\mathcal{D}|}$
    \STATE Perform instance-wise backward propagations and get $\{g_i\}_{i=1}^{|\mathcal{D}|}$
    \STATE Compute $\forall i \; \tilde{g}_i = g_i - \bar{g} $ where $\bar{g} = \frac{1}{|\mathcal{D}|}\sum_{i=1}^{|\mathcal{D}|}g_i$ 
    \STATE Compute distance matrix $M$ by Eq.~\eqref{eq:distance_matrix}
    \STATE Perform clustering $\{s_i\} \leftarrow DBSCAN(M)$, where $s_i=-1$ indicates outliers
    \STATE Update $\mathcal{D}$ by removing outliers
    \ENDIF
     \STATE Sample a mini-batch samples $\mathcal{B} \subset \mathcal{D}$  
     \STATE Perform forward passes and get $\{\ell(\theta_t;z_i)\}_{i=1}^{|\mathcal{B}|}$
    \STATE Perform instance-wise backward propagations and get $\{g_i\}_{i=1}^{|\mathcal{B}|}$
    \STATE Obtain $\nabla_\theta \mathcal{F}(\theta_t)$ by calculating the mean of the gradients whose losses are top-$|\mathcal{B}|\alpha$ largest
    \STATE Update $\{v_{t,i}\}_{i=1}^B$ according to Eq.~\eqref{eq:gradient_similarity}
    \STATE Update  $\{w_{t,i}\}_{i=1}^B$ according to Eq.~\eqref{eq:per_sample_weight}
    \STATE Update $\theta_{t+1} = \theta_t - \sum_{i=1}^B w_{t,i} g_i$
\ENDFOR
\end{algorithmic}
\end{algorithm}

\begin{table*}[t]
\caption{\label{tbl:specific_attributes} Test performance comparison on four benchmark datasets. The first block of methods are learning w/o any specified attributes but with the group ratio $\alpha$, where the best results are marked in bold, and the second best are underlined. The last second block method, i.e., MMPF, explicitly uses actual sensitive attributes during training. We omit the standard variances of $\Delta$.}
\centering
\renewcommand{\arraystretch}{1.1}
  \setlength{\tabcolsep}{1.1mm}{	
  \scalebox{0.82}{
\begin{tabular}{c|ccc|ccc|ccc|ccc}
\toprule[1.3pt]
\multirow{2}{*}{Method}  &\multicolumn{3}{c|}{UCI Adult ($\alpha=4.78\%$)}  & \multicolumn{3}{c|}{Law School ($\alpha =2.74\%$ )} & \multicolumn{3}{c|}{COMPAS ($\alpha=9.46\%$)} & \multicolumn{3}{c}{CelebA ($\alpha=7.23\%$)}\\  & ACC $\uparrow$ & WACC $\uparrow$ & $\Delta$ $\downarrow$ & ACC $\uparrow$  & WACC $\uparrow$  & $\Delta$ $\downarrow$  & ACC $\uparrow$ & WACC $\uparrow$ & $\Delta$ $\downarrow$ & ACC $\uparrow$ & WACC $\uparrow$ & $\Delta$ $\downarrow$\\
\midrule
DRO~\cite{hashimoto2018fairness}
& 0.7630$\pm$0.0726 & 0.6965$\pm$0.0676 & 0.0665
& 0.7989$\pm$0.1791 & 0.5613$\pm$0.1140 & 0.2373
& 0.5492$\pm$0.0253 & 0.4770$\pm$0.0556 & 0.0722
& 0.5219$\pm$0.1219 & 0.4586$\pm$0.0945 & 0.0633\\
CVaR~\cite{duchi2021learning}
& 0.7511$\pm$0.0269 & 0.6992$\pm$0.0292 & 0.0519
& 0.7503$\pm$0.0501 & 0.5708$\pm$0.0337 & 0.1795
& 0.5343$\pm$0.0201 & 0.4875$\pm$0.0325 & 0.0468
& 0.6903$\pm$0.0093 & 0.6475$\pm$0.0598 & 0.0428\\
BPF~\cite{martinez2021blind}
& \underline{0.8206}$\pm$0.0072 & \underline{0.7692}$\pm$0.0052 & \underline{0.0514}
& \underline{0.8698}$\pm$0.0024 & \underline{0.7592}$\pm$0.0122 & \underline{0.1106}
& \underline{0.5989}$\pm$0.0144 & \underline{0.5641}$\pm$0.0241 & \underline{0.0348}
& \textbf{0.9287}$\pm$0.0273 &\underline{0.9031}$\pm$0.0135 & \underline{0.0256}\\
IRW (Ours)
& \textbf{0.8438}$\pm$0.0074 & \textbf{0.7980}$\pm$0.0105 &\textbf{0.0458}
& \textbf{0.8715}$\pm$0.0012 & \textbf{0.7736}$\pm$0.0082 &\textbf{0.0979}
& \textbf{0.6628}$\pm$0.0068 & \textbf{0.6457}$\pm$0.0187 &\textbf{0.0171}
& \underline{0.9243}$\pm$0.0057 &\textbf{0.9109}$\pm$0.0027 & \textbf{0.0134}\\
\midrule
MMPF~\cite{martinez2020minimax}
& 0.8533$\pm$0.0192 & 0.8123$\pm$0.0920 & 0.0410
& 0.8832$\pm$0.0142 & 0.7870$\pm$0.0124 & 0.0962
& 0.6639$\pm$0.0025 & 0.6390$\pm$ 0.0145 & 0.0248
& 0.9291$\pm$0.0048 & 0.9158$\pm$0.0052 & 0.0133\\
\bottomrule[1.3pt]
\end{tabular}}}
\end{table*}

\section{Experiment}
\textbf{Dataset.} We consider four benchmark datasets that encompass three binary and one multi-class classification task. (i)~UCI Adult~\cite{asuncion2007uci}: Predicting whether an individual's annual income is above 50,000 USD. (ii)~Law School~\cite{wightman1998lsac}: Predicting the success of bar exam candidates. (iii)~COMPAS~\cite{barenstein2019propublica}: Predicting recidivism for each convicted individual. (iv)~CelebA~\cite{liu2015deep}: Predict the hair color for face images. 

\textbf{Metrics.} During the test process, we use provided sensitive attributes to divide test set into disjoint groups. Since the worst-case fairness inherits John Rawls' equal utility principle, following the research~\cite{zhai2021doro}, we evaluate the performance of trained models based on their overall accuracy (ACC) and worst-off group accuracy (WACC), where the model fairness can be expressed as the gap between ACC and WACC, that is 
\begin{equation}
\Delta = |\text{ACC}-\text{WACC}|. \nonumber
\end{equation}
Apparently, $\Delta=0$ indicates ideally fair model. We will also point out the potential weaknesses of employing such accuracy-based metrics in Section~\ref{exp:risk_view}. 

\subsection{Comparison on Specific Attributes}
This group of experiments follow the convention of recent fairness studies~\cite{martinez2021blind,chai2022self} which employ specific sensitive attributes for fairness evaluation; gender and race are selected for dataset (i-iii) and gender and young are selected for CelebA. In the context of $\alpha$-sized worst-case fairness, we will pre-compute the corresponding minimal group ratio $\alpha$ (the minimum ratio among four groups, as shown in the header row of Table~\ref{tbl:specific_attributes}) for each dataset. This pre-computed ratio will be used for model training. We take the following methods (i-iii) which are direct baselines for solving $\alpha$-sized worst-case fairness problem, while (iv) is used as a reference which explicitly uses sensitive attributed during training.(i)~DRO~\cite{hashimoto2018fairness}: This method derives a convex surrogate loss which upper bounds the empirical risk of the worst-off group. Please see the details back to Section~\ref{sec:group_proportion_bounding}. (ii)~CVaR~\cite{duchi2021learning}: It is known as the Conditional Value at Risk of level $\alpha$, which is equivalent to minimizing the $\alpha$-fraction of training points that have the highest loss~\cite{liu2021just}. (iii)~BPF~\cite{martinez2021blind}: Blind Pareto Fairness optimizes sample weights using projected gradient descent in an adversarial manner (See Appendix B.1 of~\cite{chen2017robust}), where samples with higher loss will be assigned greater weights during the training process. (iv)~MMPF~\cite{martinez2020minimax}: MiniMax Pareto Fairness is an representative attributes-aware method which explicitly uses sensitive attributes to split training data and encourages the equal model accuracy across groups via solving a multi-objective problem. Each experiment will be repeated for 10 times, and the average and standard deviation are reported.
\begin{figure*}
\centering
  \begin{minipage}[b]{0.245\textwidth}
    \centering
    \includegraphics[width=\textwidth]{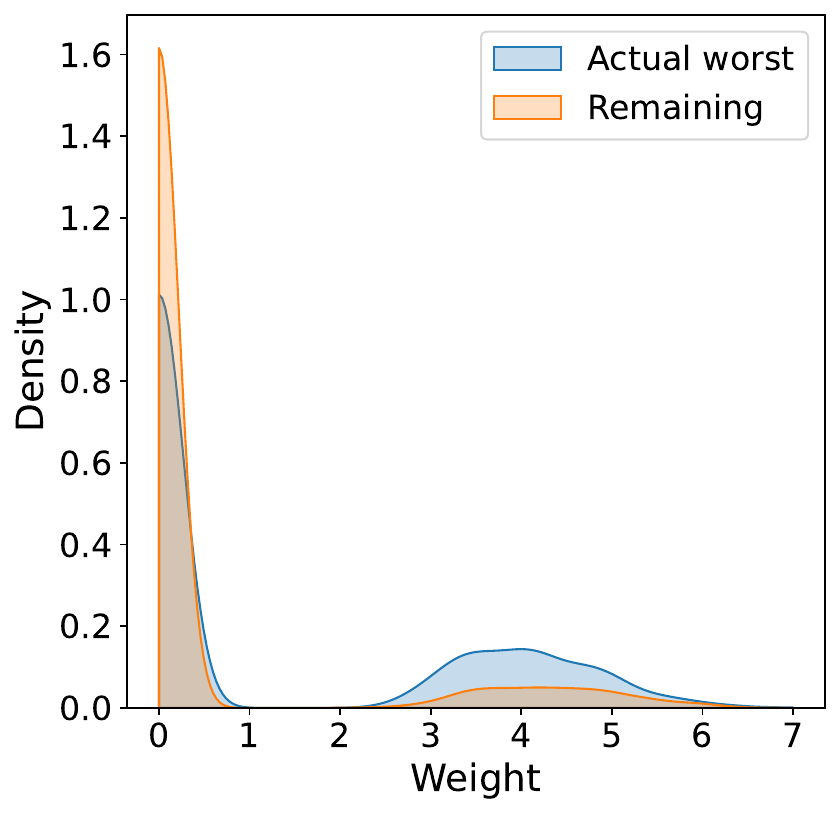}
    \centerline{(a) UCI Adult }
  \end{minipage}
  \begin{minipage}[b]{0.245\textwidth}
    \centering
    \includegraphics[width=\textwidth]{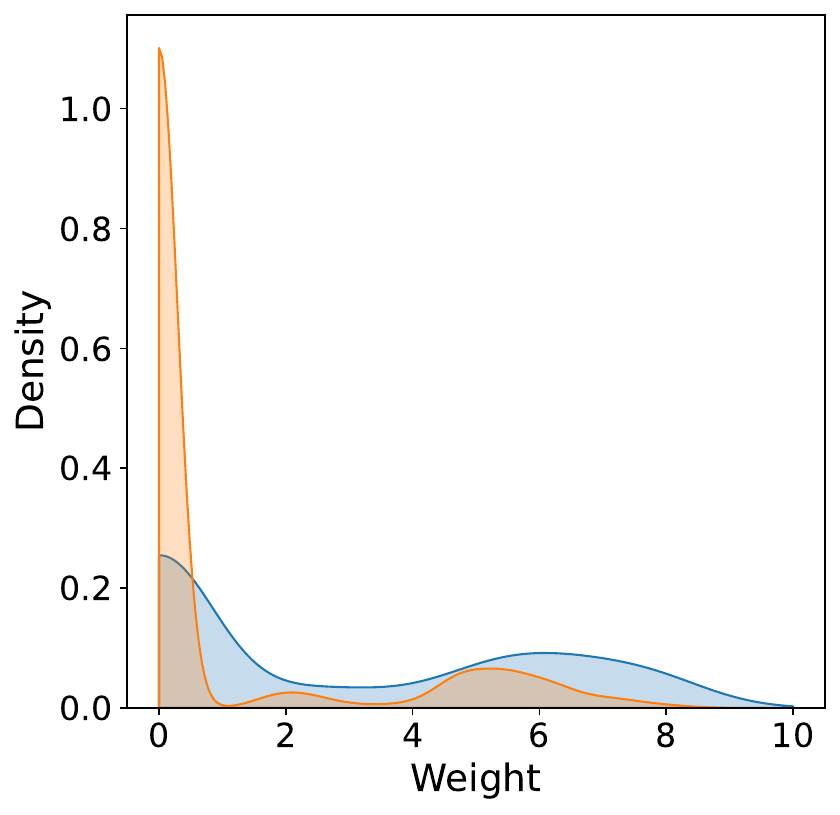}
    \centerline{(b) Law School}
  \end{minipage}
  \begin{minipage}[b]{0.245\textwidth}
    \centering
    \includegraphics[width=\textwidth]{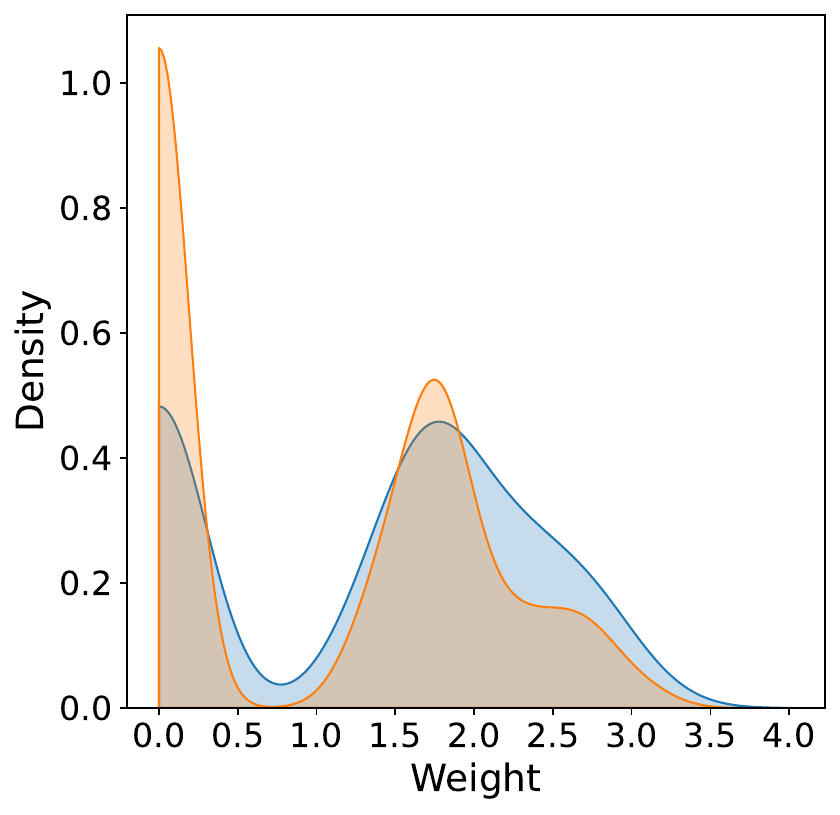}
    \centerline{(c) COMPAS}
  \end{minipage} 
 \begin{minipage}[b]{0.242\textwidth}
    \centering
    \includegraphics[width=\textwidth]{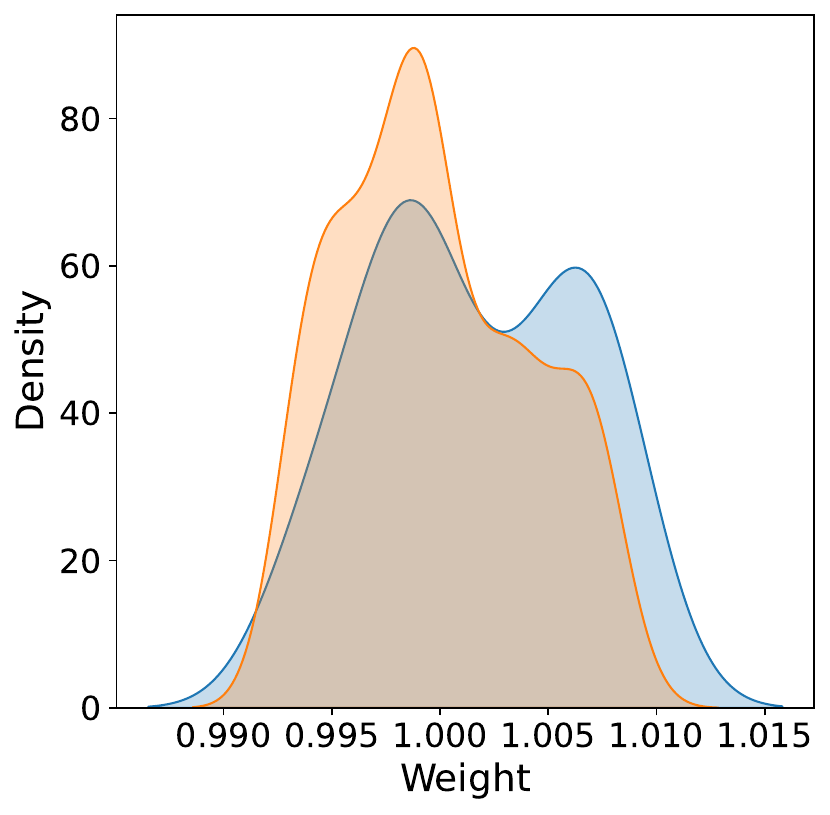}
    \centerline{(d) CelebA}
  \end{minipage}  
  \caption{ Weight distribution of training samples for our IRW. For each dataset, we record the weights of training samples at the first training epoch and use kernel density estimation to plot their distribution. The actual worst indicates the group which has the lowest accuracy on test set, and all rest are included in the remaining.
   \label{fig:weight_distribution}}\vskip-0.1in
\end{figure*}

\begin{figure*}
\centering
  \begin{minipage}[b]{0.245\textwidth}
    \centering
    \includegraphics[width=\textwidth]{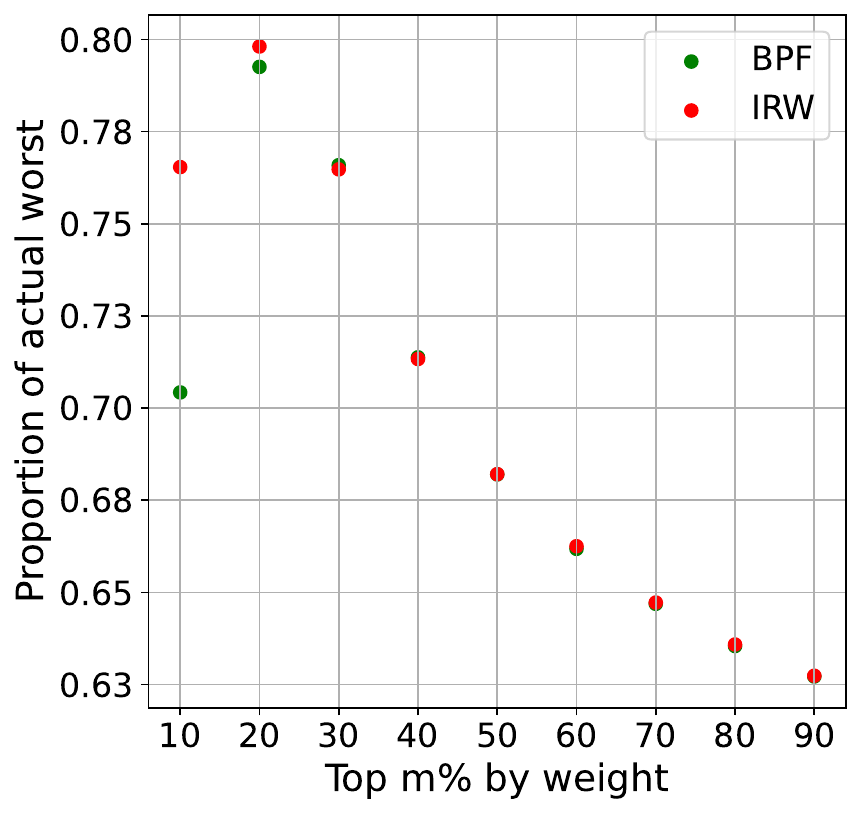}
    \centerline{(a) UCI Adult }
  \end{minipage}
  \begin{minipage}[b]{0.245\textwidth}
    \centering
    \includegraphics[width=\textwidth]{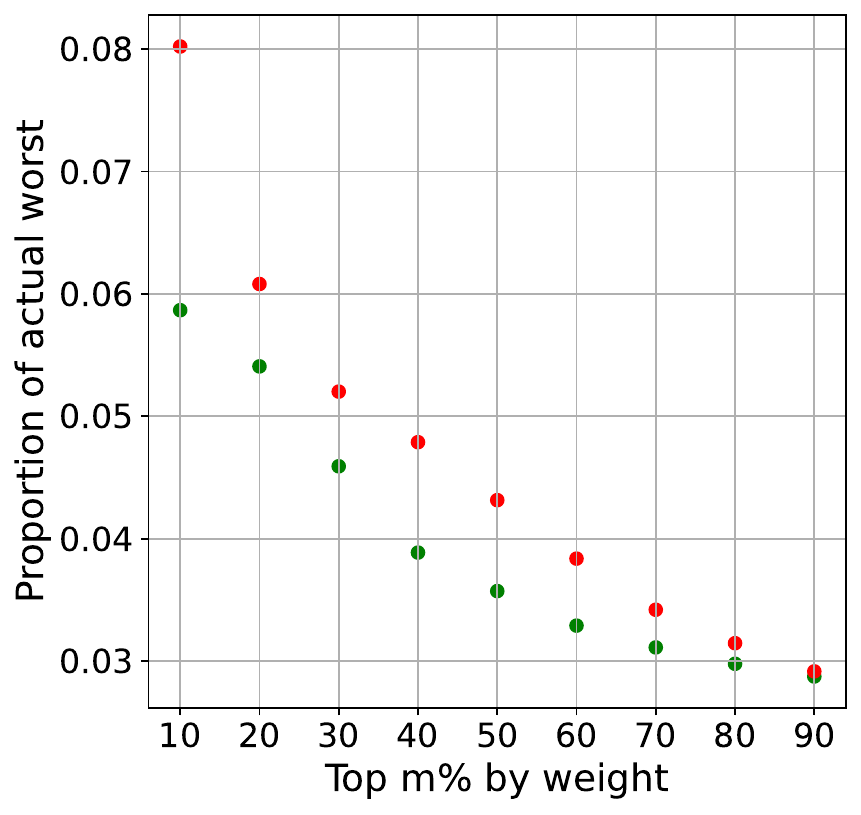}
    \centerline{(b) Law School}
  \end{minipage}
  \begin{minipage}[b]{0.245\textwidth}
    \centering
    \includegraphics[width=\textwidth]{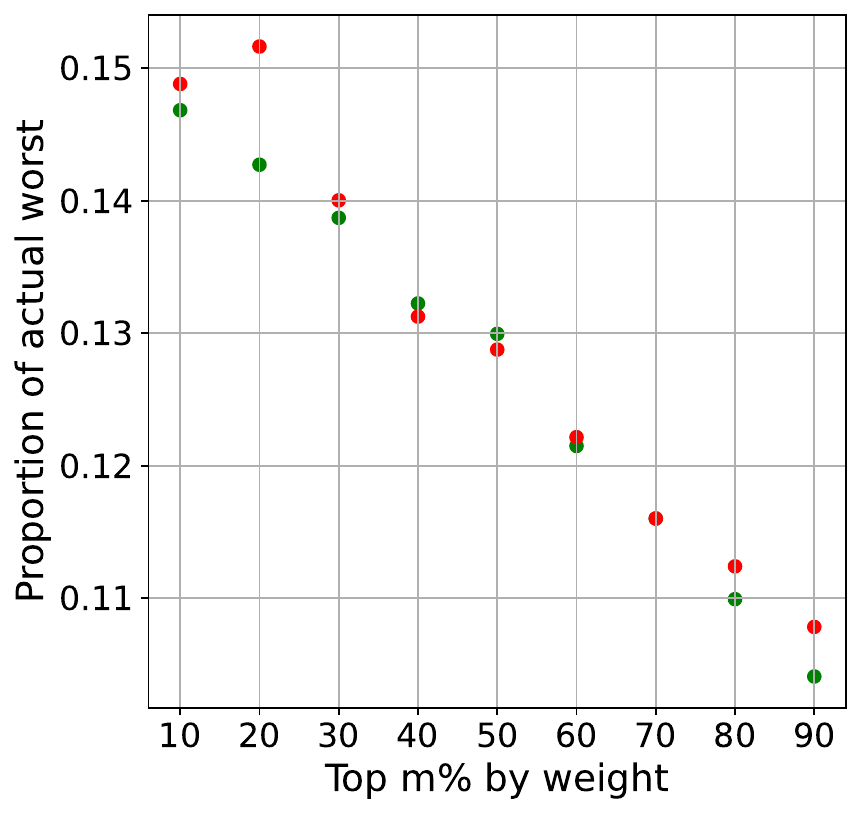}
    \centerline{(c) COMPAS}
  \end{minipage} 
 \begin{minipage}[b]{0.242\textwidth}
    \centering
    \includegraphics[width=\textwidth]{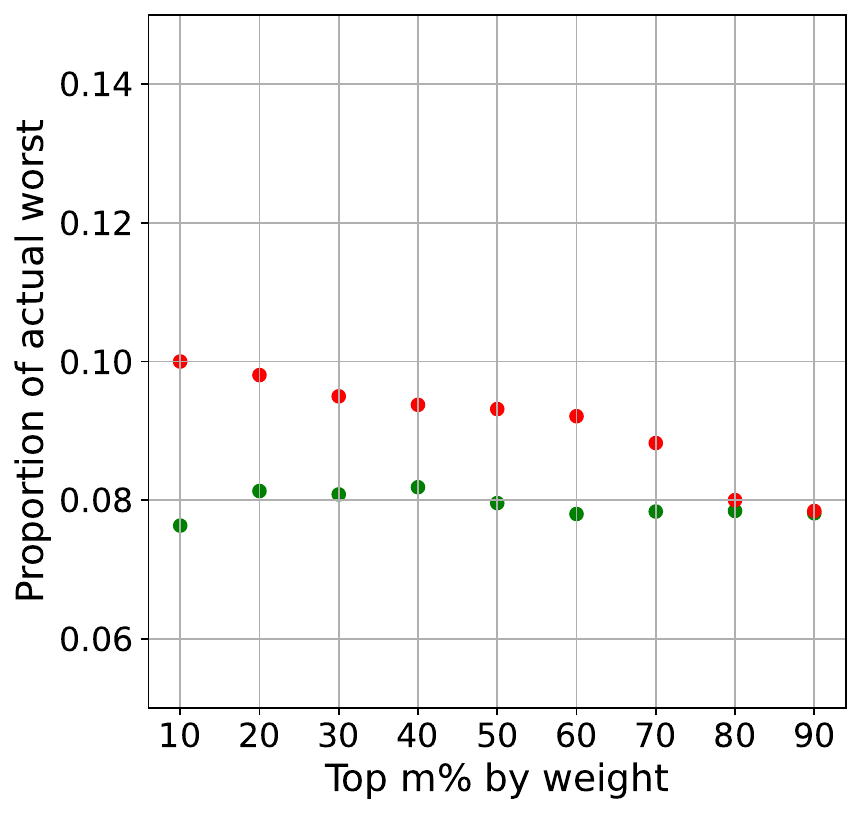}
    \centerline{(d) CelebA}
  \end{minipage}  
  \caption{ The proportion of actual worst-off group among top $m\% (m=10,20,...,90)$ training samples selected by reassigned weights. We compare BPF and our IRW method on four datasets.\label{fig:top_proportion}}\vskip-0.1in
\end{figure*}

\begin{figure*}[t]
\centering
  \begin{minipage}[b]{0.24\textwidth}
    \centering
    \includegraphics[width=\textwidth]{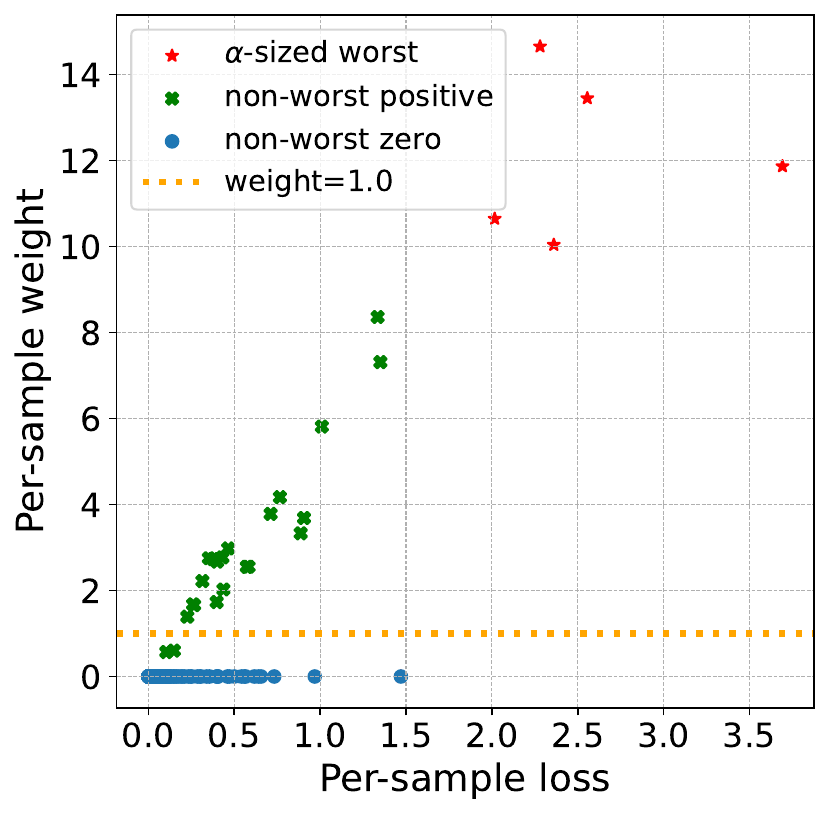}
    \centerline{(a) UCI Adult }
  \end{minipage}
  \begin{minipage}[b]{0.248\textwidth}
    \centering
    \includegraphics[width=\textwidth]{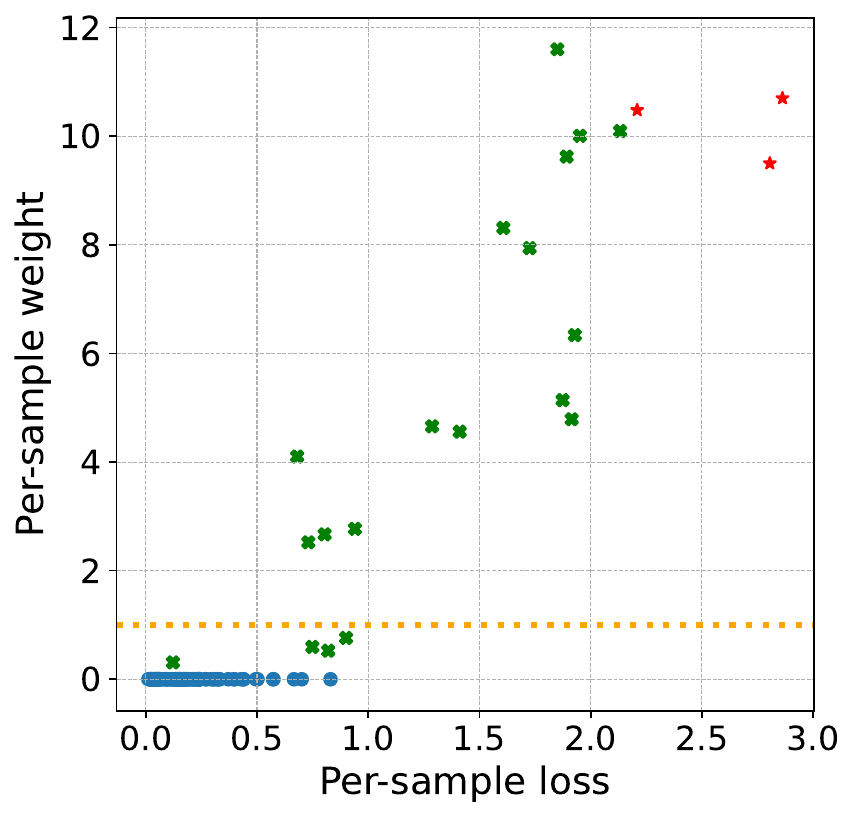}
    \centerline{(b) Law School}
  \end{minipage}
  \begin{minipage}[b]{0.235\textwidth}
    \centering
    \includegraphics[width=\textwidth]{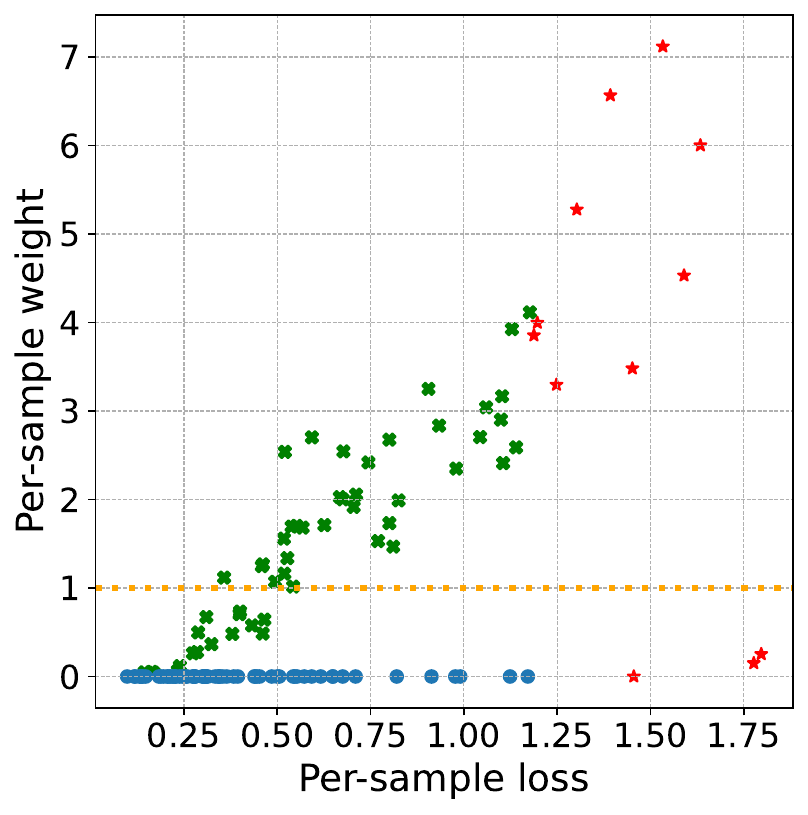}
    \centerline{(c) COMPAS}
  \end{minipage} 
 \begin{minipage}[b]{0.244\textwidth}
    \centering
    \includegraphics[width=\textwidth]{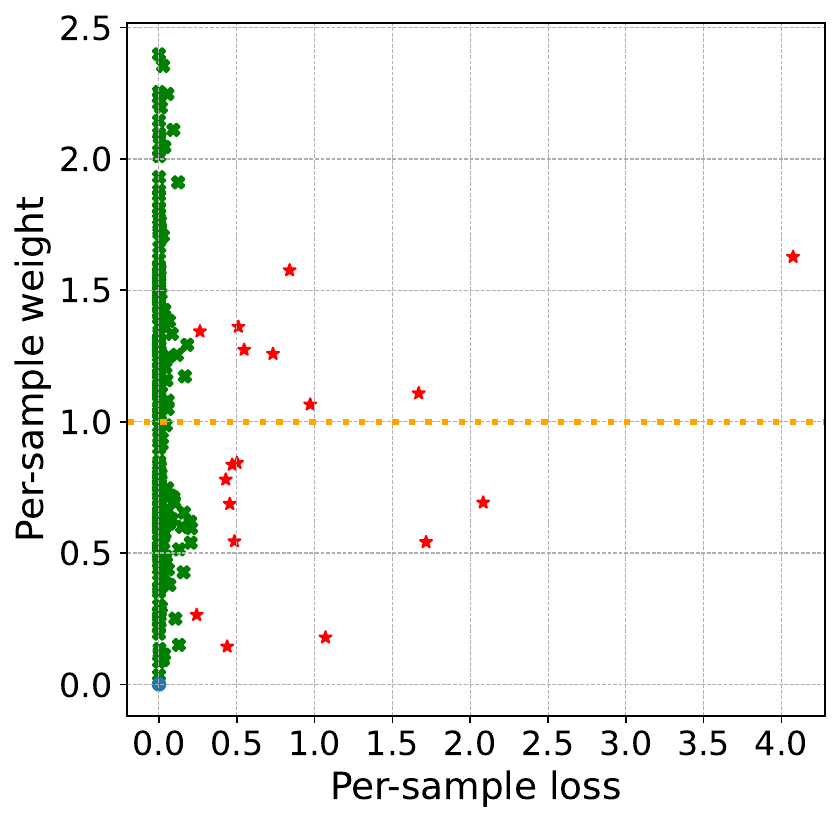}
    \centerline{(d) CelebA}
  \end{minipage}  
  \caption{ Per-sample weight versus their loss values on a randomly selected batch at the end of training process on four datasets. The number of $\alpha$-sized worst samples (red stars) is $\text{int}(b\times\alpha)$, where $b=128$ for (a)-(c) and $b=256$ for (d), and $\alpha$ is indicated by the head row of Table~\ref{tbl:specific_attributes}. The non-worst positive-weighted samples (green crosses) and non-worst zero-weighted samples (blue points) are the ones whose gradients yield positive or zero $w_i$ at the iteration $t$, respectively. (Refer to Eq.~\eqref{eq:per_sample_weight}).  \label{fig:weight_loss}}\vskip-0.1in
\end{figure*}

Table~\ref{tbl:specific_attributes} shows test ACC, WACC, and $\Delta$ of each method on four datasets. We can see that: (1)~Our IRW method achieves the best performance across almost all datasets, demonstrating the superiority of leveraging individual gradient to calculate sample importance. Also, IRW achieves comparable or slightly worse performance than MMPF, which has utilized sensitive attributes during training. 
(2)~BPF most often achieves the second-best performance, demonstrating that adversarial reweighing strategy is effective in achieving $\alpha$-sized fairness. It is worthwhile to point out that BPF involves a few of hyper-parameters and also heuristic controls\footnote{Refer to their officially implemented code repository via \url{https://github.com/natalialmg/BlindParetoFairness}.}, and we applied their default setup throughout the experiments. (3)~CVaR uses only the $\alpha$-fraction of data for model training, resulting in a significant drop in ACC and WACC, as well as in fairness results, compared to BPF and our IRW. (4)~DRO usually achieves the worst performance expect that it is slightly better on accuracy than CVaR for the COMPAS dataset, which is attributed to the commonly smaller $\alpha$ across all datasets (also observed in experiments of~\cite{martinez2021blind}). Because a smaller $\alpha$ indicates a larger radius of the chi-squared ball in Eq.~\eqref{eq:distributionl_robust_chi_square_ball}, leading to an unavoidable loss in model accuracy. Such unsatisfying model utility makes DRO impractical to use, for example, on the CelebA dataset.

\subsection{IRW Works Better than BPF}
We first verify the efficacy of IRW by examining how samples from the actual worst-off group are treated during training, given only a group ratio prior. To do this, we monitor the individual weights assigned to all training samples. To align with the standard training process, we keep a base weight of 1, with weights higher or lower than 1 being increased or decreased accordingly. Other subsequent presentations regarding to IRW follow this setup. Fig.~\ref{fig:weight_distribution} depicts the weight distribution of training samples by splitting all training samples into the \emph{actual worst}-off group and \emph{remaining} samples. Across all datasets, we observe that while the weight distribution does not form two distinct peaks, as would be ideal, the actual worst-off group samples tend to have a relatively higher density in the range of larger weights. This indicates that IRW implicitly upweights the actual worst-off group samples, even though sensitive attributes are not accessible during model training. This observation evidenced the efficacy of our IRW method.

Although can be understood by Eq.~\ref{eq:adv_weight_obj}, BPF practically uses the samples weights which do not lies on a simplex. Normalizing them may distort the weight density. To further exploit why our method IRW can outperform BPF, we compare them using the following approach. We select the top $m\%$ training samples based on their corresponding weights, and compute the proportion of actual worst-off group samples within these selected samples. Since we expect that the actual worst-off group can be assigned with higher weights, a higher proportion indicates a fairer model. The results are shown as Fig.~\ref{fig:top_proportion}, from which we observe that at almost every $m$ across four datasets, our IRW achieves the higher proportion value than BPF, demonstrating that our method can better leverage prior to implicitly encourage the fitting for the actual worst-off groups. It is also worthwhile to notice that the proportion on the UCI Adult dataset is not aligned with the prior when $m$ approaches to $100$ while other three do. This is because the minority group sometimes is not necessarily the underrepresented group. Although loosing the upper bound in Eq.~\eqref{eq:bounded_by_top_N_alpha}, it shows a distinction assumption that minority is always underrepresented~\cite{liu2021just}.

\subsection{Per-sample Reweight Mechanism }\label{sec:per-sample_reweight_mechanism}
Figs.~\ref{fig:weight_distribution} and~\ref{fig:top_proportion} demonstrate that IRW increases the weights of training samples from the actual worst-off group. To delve deeper, we analyze how these weights correlate with their loss values, which is central to the per-sample weight mechanism for fairness. To this end, we randomly select a batch of samples at the end of the training phase on four datasets, where the models are assumed to be close to convergence, with both weight and loss values remaining stable afterwards.

Fig.~\ref{fig:weight_loss} shows the experimental results of per-sample weight versus loss, based on which we have following observations. (1)~IRW typically assigns greater weights to samples with greater losses, aligning with the idea that poorly-fit samples are more likely to come from underrepresented groups. (2)~The weights of $\alpha$-sized worst samples are not necessarily larger than those of other samples with relatively smaller, as observed on last three datasets, which thus differs from CVaR. This occurs because the weight of an $\alpha$-sized worst sample may be smaller if its gradient is not consistent with the combined gradient, as indicated by Eq.~\eqref{eq:gradient_similarity}. (3)~IRW can produce sparse weights, with zero weights often assigned to samples with smaller losses. This property eliminates the need to manually enforce a minimum update, as required in BPF, which typically demands additional effort to identify. (4)~The weight-loss correlation on CelebA stands out from the others, where many training samples with very small losses have increased weights, while some of the $\alpha$-sized worst samples have decreased weights. This suggests that, in our IRW approach, training samples contributing to fairness from a gradient perspective will be widely identified, even if their losses are negligible.


\begin{table}[t]
\caption{\label{tbl:f1_uci_adult} Test performance comparison on the UCI Adult dataset where specific sensitive attributes are used for splitting test set into different groups. The best results are marked as bold, and the second best are underlined.}
\centering
\renewcommand{\arraystretch}{1.1}
  \setlength{\tabcolsep}{1.1mm}{	
  \scalebox{1}{
\begin{tabular}{c|ccc}
\toprule[1.3pt]
Method  & F1 $\uparrow$ & WF1 $\uparrow$ & $\Delta_{\text{F1}}$ $\downarrow$ \\
\midrule
DRO
& 0.3796$\pm$0.0726 & 0.1581$\pm$0.0447 & 0.2215\\
CVaR
& 0.3999$\pm$0.1311 & 0.2231$\pm$0.0945 & 0.1768\\
BPF
& \underline{0.4841}$\pm$0.0517 & \underline{0.4079}$\pm$0.0631 & \underline{0.0762}\\
IRW
& \textbf{0.5749}$\pm$0.0493 & \textbf{0.5024}$\pm$0.0502 & \textbf{0.0725}\\
\bottomrule[1.3pt]
\end{tabular}}}
\end{table}

\subsection{Evaluation Metric Matters}
\emph{Class imbalance.} F-score is particularly used in binary classification tasks where one class may be significantly underrepresented. We realize that the UCI Adult dataset tends to be dominated by true negative samples due to its imbalanced nature, with a positive to negative ratio of 1:4. Without changing each training method, we evaluate their test performance by utilizing the F-score metric. The results are shown as Table~\ref{tbl:f1_uci_adult}. It is evident that with F1 being the classification metric, IRW consistently achieves the best performance, demonstrating that despite originating from group accuracy, i.e., Eq.~\eqref{eq:utility_as_acc}, our method IRW is not biased towards accuracy metric. Furthermore, we also observe that the performance difference between IRW and BPF is larger in terms of F1 measurement, which justifies the advantage of leveraging per-sample gradient information in our reweighing strategy. 
\begin{figure}[t]
\centering
  \begin{minipage}[b]{0.24\textwidth}
    \centering
    \includegraphics[width=\textwidth]{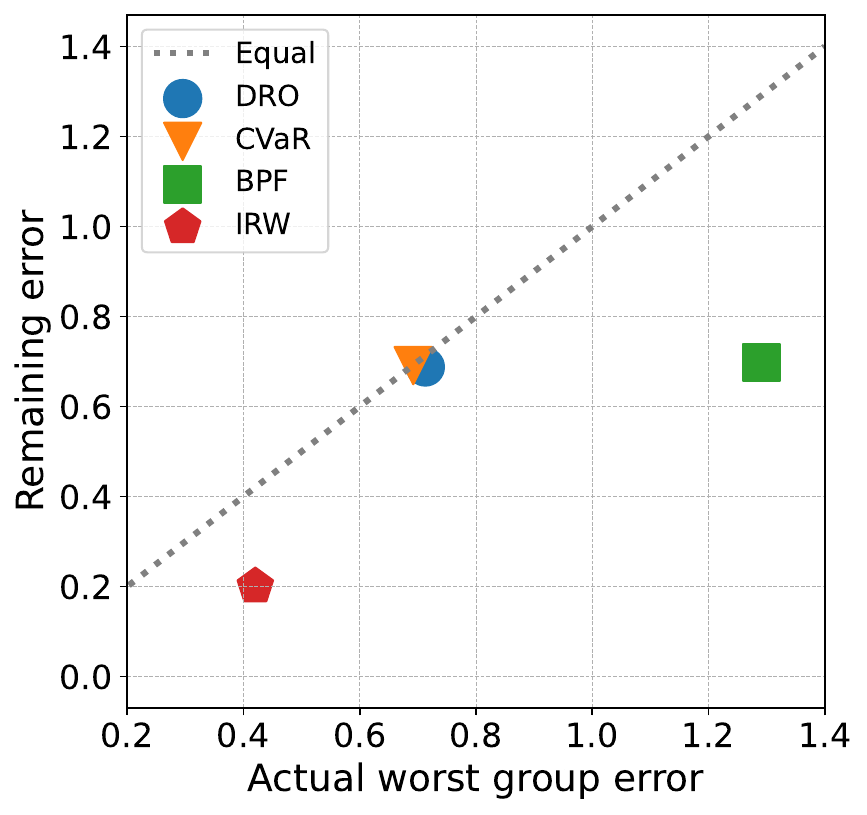}
    \centerline{(a) Group-wise error}
  \end{minipage}
  \begin{minipage}[b]{0.24\textwidth}
    \centering
    \includegraphics[width=\textwidth]{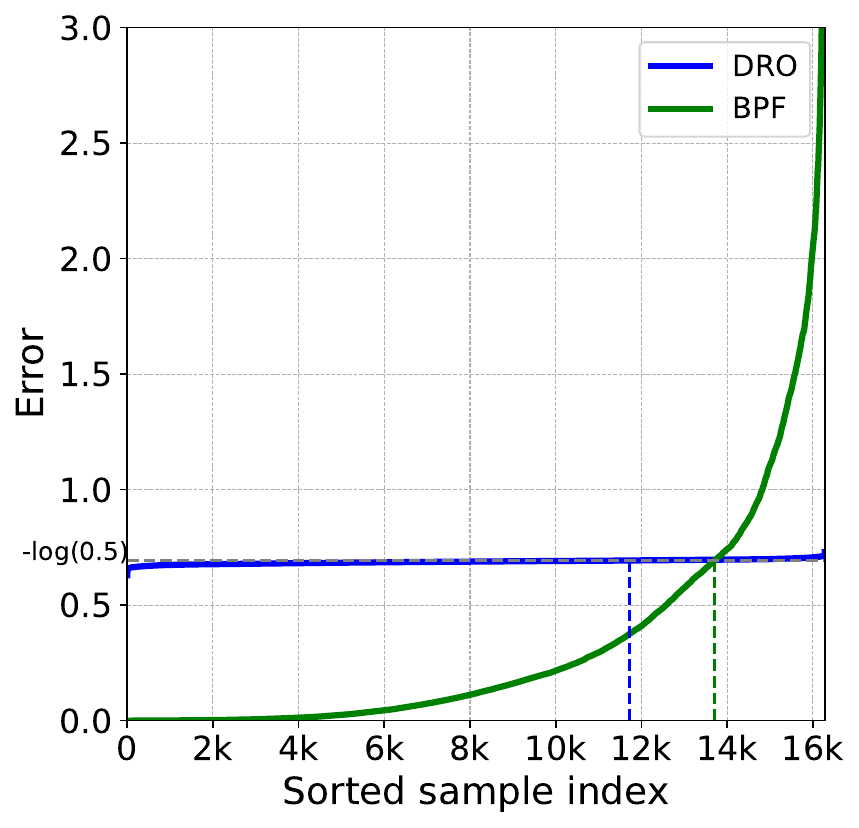}
    \centerline{(b) Instance-wise error}
  \end{minipage}
  \caption{Classification errors comparison on the UCI Adult dataset. The y-axis is truncated above 3 for better visualization. \label{fig:uci_adult_error}}\vskip-0.1in
\end{figure}

\label{exp:risk_view}
\emph{Classification error.} According to Table~\ref{tbl:specific_attributes}, we have observed that DRO achieves the smallest $\Delta$ on the UCI Adult dataset, but significantly sacrifices model accuracy. Specifically, compared to our IRW, model accuracy drops by more than 0.20 while fairness only improves by only around 0.01. To understand this trade-off, we take a closer look at their classification errors on test set. Fig.~\ref{fig:uci_adult_error}(a) shows the classification error of the actual worst-off group versus remaining samples. We observe that DRO and CVaR both achieve the approximately equal errors which however increase the overall error by comparing with IRW, and BPF is the least fair in terms of error. Such results are somehow contrary to what we have seen in Table~\ref{tbl:specific_attributes}. For example, DRO achieves a lower overall error than BPF, yet its accuracy is identified as the worst. We further explain such inconsistency on metrics by connecting prediction errors and decision making. Fig.~\ref{fig:uci_adult_error}(b) shows the instance-wise errors, with the x-axis representing the test sample indices, sorted in an ascending order of their errors. The y-axis is truncated above 3 for improved clarity in presentation. Since this is a binary classification on the Adult dataset, we adopt the reference error of $-\log(0.5)$ as a decision boundary (indicated by dashed vertical lines) to categorize all test samples. Samples with an error below this value are considered correctly classified, while those above it are regarded as misclassified. We observe that although the errors of misclassified samples of BPF can be large, the overall accuracy of BPF remarkably outperforms DRO, as the dashed line for BPF is substantially to  the right of the dashed line for DRO. It is also worthwhile to note that the performance gap between DRO and BPF in Fig.~\ref{fig:uci_adult_error}(b) is smaller than the reported average value in Table~\ref{tbl:specific_attributes}, which is attributed to the larger variance of DRO.

\subsection{Global vs. Local Worst-Case Fairness Training}\label{sec:globla_vs_local}
\begin{figure}[t]
\centering
  \begin{minipage}[b]{0.235\textwidth}
    \centering
    \includegraphics[width=\textwidth]{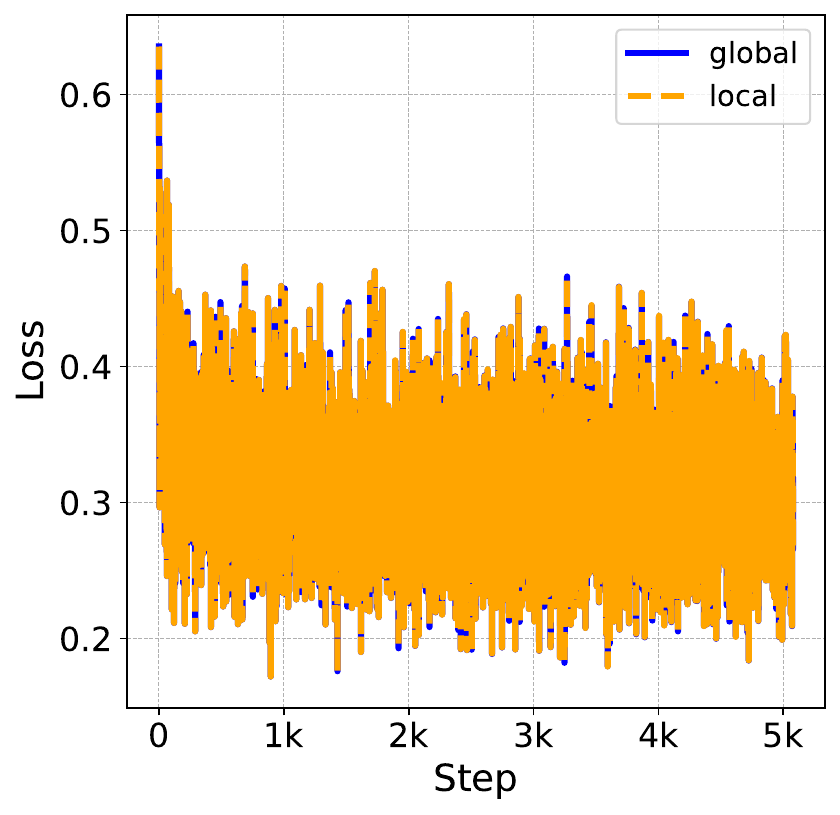}
    \centerline{(a) Training loss}
  \end{minipage}
  \begin{minipage}[b]{0.24\textwidth}
    \centering
    \includegraphics[width=\textwidth]{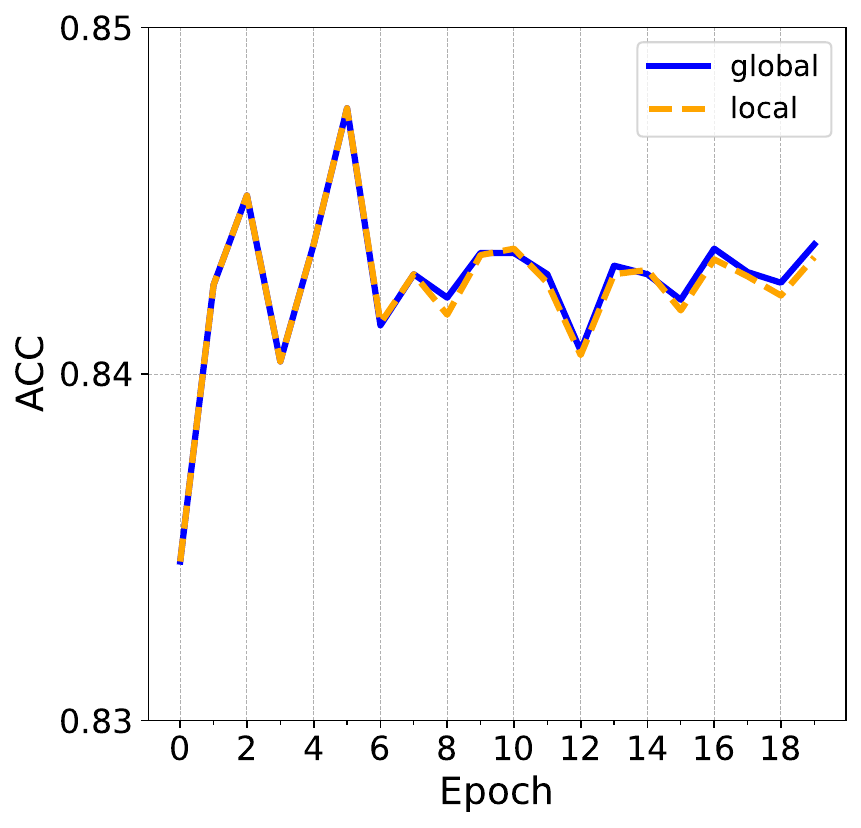}
    \centerline{(b) Test accuracy}
  \end{minipage}
  \caption{Comparison between global and local worst-case curves on the UCI Adult dataset, where (a) is the loss versus each training step and (b) is test accuracy on each epoch. \label{fig:uci_adult_train_loss_and_test_acc}}\vskip-0.1in
\end{figure}

We compare the efficiency of global and local worst-case training with the Adult dataset as an example, corresponding to schemes~(i) and~(ii) in Section~\ref{sec:stochastic_update}. Their training loss curves and overall test accuracy are exhibited as Fig.~\ref{fig:uci_adult_train_loss_and_test_acc}. The results show that local worst-case training which selects $\alpha$-sized worst samples among each mini-batch are aligned with the global counterpart; their training loss and test accuracy curves are close to each other, with a very tiny discrepancy after 7th epoch from Fig.~\ref{fig:uci_adult_train_loss_and_test_acc}(b). Apart from this conclusion, we observe that their loss curves fluctuate significantly within the range of 0.2 to 0.4, which is essentially shared by all worst-case reweighting strategy, as also noted by~\cite{zhai2021doro}. Similarly, their test accuracy fluctuates across epochs, but within a relatively smaller range of around 0.5\%. Additionally, we record the wall-clock time for training; global training takes around 31 seconds while local training takes 42 seconds. By tracking the time consumption of each component during local training, we find that multiple graph-retention steps (e.g., \texttt{retain\_graph=true} in Python) increase the overall time cost, a process that is not required in global training. However, local training remains an acceptable option when full-batch forward passes are not feasible in some scenarios, e.g., limited computation resources. 

\subsection{Robustness Against Outliers}\label{sec:robustness_to_outliers}
We compare IRWO with three fairness methods ARL~\cite{lahoti2020fairness}, DORO~\cite{zhai2021doro}, and GraSP~\cite{zeng2022outlier}, which have been used to address outliers for worst-case modeling, as discussed in Section~\ref{sec:outliers_removal}. For methods with specific hyper-parameters, we use their default setup to report the final results. The COMPAS and CelebA datasets are used in this group of experiments because they two are considered to contain outliers~\cite{zhai2021doro}. 

Table~\ref{tbl:outlier} summarizes the results of all robust worst-case fairness methods, from which we have the following observations. (1)~IRWO achieves the best performance among all the methods compared, showing the superiority of the proposed method. (2)~Referring to the results in Table~\ref{tbl:specific_attributes}, we find that the variants with outliers removal improve the standard counterparts; DORO significantly improves DRO and IRWO also generally outperforms IRW over accuracy. (3)~ARL performs slightly worse than our IRWO, which can be attributed to the its formulation that does not utilize $\alpha$. (4)~GraSP uses the same outlier removal strategy as ours but underperforms due to its reliance on estimated sensitive attributes from clustering results.

\begin{table*}[h]
\caption{\label{tbl:outlier} Test performance comparison on four benchmark datasets where specific sensitive attributes are used for splitting test set into different groups. The best results are marked as bold, and the second best are underlined.}
\centering
\renewcommand{\arraystretch}{1.1}
  \setlength{\tabcolsep}{1.1mm}{	
  \scalebox{1}{
\begin{tabular}{c|ccc|ccc}
\toprule[1.3pt]
\multirow{2}{*}{Method} & \multicolumn{3}{c|}{COMPAS} & \multicolumn{3}{c}{CelebA }\\  & ACC $\uparrow$ & WACC $\uparrow$ & $\Delta$ $\downarrow$ & ACC $\uparrow$  & WACC $\uparrow$  & $\Delta$ $\downarrow$ \\
\midrule
ARL~\cite{lahoti2020fairness}
& \underline{0.6612}$\pm$0.0091 & \underline{0.6255}$\pm$0.0188 & \underline{0.0357}
& 0.9058$\pm$0.1038 & 0.8689$\pm$0.1004 & 0.0369\\
DORO~\cite{zhai2021doro}
& 0.5719$\pm$0.0241 & 0.5241$\pm$0.0427 & 0.0478
& 0.7714$\pm$0.0333 & 0.7350$\pm$0.0400 & 0.0364\\
GraSP~\cite{zeng2022outlier}
& 0.6009$\pm$0.0211 & 0.5592$\pm$0.0481 & 0.0417
& \underline{0.9255}$\pm$0.0501 & \underline{0.8969}$\pm$0.0337 & \underline{0.0286}\\
IRWO (Ours)
& \textbf{0.6688}$\pm$0.0113 & \textbf{0.6388}$\pm$0.0213 & \textbf{0.0300}
&\textbf{0.9307}$\pm$0.0057 &\textbf{0.9209}$\pm$0.0241 & \textbf{0.0098}\\
\bottomrule[1.3pt]
\end{tabular}}}
\end{table*}

\section{Discussion}
Beyond loss values, we have found that individual gradients provide a complementary view for characterizing a sample's importance to the fairness objective. However, gradients are sometimes too complex to work with, especially in larger neural networks involving millions of billions of parameters. In such cases, the gradient representation resides in a high-dimensional space, posing challenges for storage and computation. One workaround we apply to the experiments on CelebA dataset is selecting the last few layers, which are more closely related to label information from the perspective of information bottleneck~\cite{kawaguchi2023does}. We emphasize that the effective selection of key layers for specific tasks remains an open question, while some potential approaches can be considered, such as fisher information~\cite{soen2021variance} or layerwise search~\cite{li2023earning}. 


Resampling has been a parallel way to achieve fairness~\cite{zhang2020fairness,romano2020achieving}. Despite various resampling  motivations applied in these works, we are interested in its technical connection to reweighting in the context of $\alpha$-sized worst-case fairness objective. Very fundamentally, we question whether the weights derived from IRW can be treated as the probability of resampling during the training process. We provide a corollary based on the recent work in group-imbalance learning~\cite{DBLP:conf/iclr/AnYZ21}, which claims that resampling always exhibits lower variance than the corresponding reweighting. While this corollary holds theoretically only when stochastic gradient descent is used (Please refer to Appendix for a formal presentation), we will explore this direction in future work.

\section{Conclusion}
This paper introduces ``$\alpha$-sized worst-case fairness'' problem where only the the minimal group ratio $\alpha$ is known during training, but the goal is to diminish the prejudice on under-represented subgroups defined by inaccessible demographic information. We justify the use of this group ratio prior, which can be derived from random responses ensured by the well-known local differential privacy framework. Unlike existing methods that simply prioritize training on samples with relatively larger loss, we propose an intrinsic reweighting strategy that assigns individual weights based on per-sample gradient information. This gradient-based approach also naturally helps in removing potential outliers, which have been identified as a significant risk in worst-case fairness formulations. Our experiments demonstrate the superiority of IRW and its variant, IRWO, in achieving fairness across four benchmark datasets, establishing a foundation of benchmark results for future research.


\appendix
\textbf{Proposition 1.} \emph{Given the minimal group proportion $\alpha$, we have $\mathcal{J}_{wg}(\theta) \le \mathcal{J}_{N\alpha}(\theta;\alpha) \le \mathcal{J}_{dr}(\theta;\alpha)$ for all $\theta \in \Theta$.} 
\begin{proof}
The first inequality is an obvious conclusion of Eq.~\eqref{eq:bounded_by_top_N_alpha}, and we show how the second inequality holds. Let $P_{N\alpha}$ be the distribution from which the examples belong to top $N\alpha$ largest losses can be drown. Then we have $P=\alpha P_{N\alpha}+(1-\alpha)P_{res}$, where $P_{res}$ denotes the distribution from which all the rest of examples can be drawn.
\begin{equation}\label{eq:N_alpha_chi_distance}
\begin{aligned}
   D_{{\chi}^2}(P_{N\alpha}\|P) &= \int_z\left(\frac{P_{N\alpha}(z)}{P(z)}-1\right)^2 P(z)dz\\
   & \le \int_z {\left( \frac{1}{\alpha}-1 \right)^2} P(z)dz\\
   & = r
\end{aligned}
\end{equation}
Eq.~\eqref{eq:N_alpha_chi_distance} showed that $P_{N\alpha} \in \mathcal{B}(P,r)$. Since the sup in $\mathcal{J}_{dr}(\theta;\alpha)$ is over all $Q \in \mathcal{B}(P,r)$ according to Eq.~\eqref{eq:distributionl_robust_chi_square_ball}, the upper bound follows.
\end{proof}

\noindent\textbf{Theorem 2.} \emph{If the feasible set defined by $C$ is convex, the optimal solution of Eq.~\eqref{eq:adv_weight_obj} implies that a higher loss corresponds strictly to a higher weight.}
\begin{proof}
For any two selected samples $z_i$ and $z_j$ with corresponding weights $w_i$ and $w_j$ obtained by solving Eq.~\eqref{eq:adv_weight_obj}, we prove $\ell(\theta;z_i) > \ell(\theta;z_i) \Rightarrow w_i > w_j$. Let $w=[...,w_i,...,w_j,...]$ be the optimal solution of Eq.~\eqref{eq:adv_weight_obj}, and we have following inequality
\begin{equation}\label{eq:swap_weights}
\sum_{k=1}^{N} w_k \ell(\theta;z_k) > w_j \ell(\theta;z_i) + w_i \ell(\theta;z_j) + \sum_{k \neq {i, j}}^{N} w_k \ell(\theta;z_k) ,
\end{equation}
where we swap $w_i$ and $w_j$ on the RHS to satisfy the simplex constraint. By canceling the identical terms and simplifying the Eq.~\eqref{eq:swap_weights}, we arrive at
\begin{equation}\nonumber
(w_i-w_j)\left( \ell(\theta;z_i)-\ell(\theta;z_j) \right) > 0.
\end{equation}
Then we get $w_i > w_j$, which completes the proof.
\end{proof}

\noindent \textbf{Corollary.} \emph{Supposing that reweighting and resampling takes the same sample importances, the gradient variance of resampling is not greater than reweighing if SGD is used as the optimizer.}
\begin{proof}
Let $P$ be the sampling proportion by which we get training samples. The conditional variance of reweighting under SGD can be written as 
\begin{equation}
\begin{aligned}
    \sigma_w^2(\theta)  
    &= \mathbb{E}_{z \sim P}[  \nabla_{\theta} \ell(\theta;z)]^2 - \mathbb{E}_{z \sim P}^2 [\nabla_{\theta} \ell(\theta;z)]\\
    &= \sum_{i=1}^N N(w_i)^2||g_i||^2 - ||\frac{1}{N}\sum_{i=1}^N w_ig_i||^2\\
\end{aligned}
\end{equation}
In resampling, one either repeats sampling some data points (i.e., oversampling) or removes some (i.e., undersampling) and finally gets $N'$ samples~\cite{DBLP:conf/iclr/AnYZ21}. We use $Q$ to denote the sampling probability with which each observed data point participates in training. The conditional variance of resampling under SGD as
\begin{equation}\label{eq_resampling_conditional_variance}
\begin{aligned}
    \sigma_s^2(\theta) 
    &= \mathbb{E}_{z \sim Q}[ \nabla_{\theta} \ell(\theta;z)]^2 - \mathbb{E}_{z \sim Q}^2 [\nabla_{\theta} \ell(\theta;z)]\\
    &= \frac{1}{N'}\sum_{i=1}^{N'} ||g_i||^2 - ||\frac{1}{N'}\sum_{i=1}^{N'} g_i||^2 \\
    &= \sum_{i=1}^N w_i ||g_i||^2 - ||\frac{1}{N}\sum_{i=1}^N w_ig_i||^2
\end{aligned}
\end{equation}
These formulas indicate that $\sigma_w^2 \ge \sigma_s^2$. The variance of reweighting can be much larger if the importance distribution is far away from the non-uniform distribution.
\end{proof}

\ifCLASSOPTIONcaptionsoff
  \newpage
\fi

\bibliographystyle{IEEEtran}
\bibliography{Reference}
\end{document}